\documentclass[conference]{IEEEtran}
\IEEEoverridecommandlockouts
\usepackage{cite}
\usepackage{amsmath,amssymb,amsfonts}
\usepackage{graphicx}
\usepackage{textcomp}
\usepackage{tabularray}
\usepackage{caption}
\usepackage{subfig}
\usepackage{multicol,lipsum}
\usepackage{xcolor}
\usepackage{comment}
\def\BibTeX{{\rm B\kern-.05em{\sc i\kern-.025em b}\kern-.08em
    T\kern-.1667em\lower.7ex\hbox{E}\kern-.125emX}}
\usepackage{booktabs}
\usepackage{amsmath}
\usepackage{amssymb} 
\usepackage{float}
\usepackage{graphicx}
\usepackage{svg}
\usepackage{algorithm}
\usepackage{algpseudocode}
\usepackage{caption}
\definecolor{ref}{RGB}{0,50,160}

\usepackage{algorithm,algpseudocode}

\begin{document}
\title{Will the Carbon Border Adjustment Mechanism Impact European Electricity Prices? A GNN-Based Network Analysis}

\author{Jiachen Shen, Jian Shi, Dan Wang and Zhu Han
\thanks{This work is partially supported by NSF ECCS-2338158}

\thanks{J. Shen, J. Shi, and Z. Han are with the Departments of Electrical and Computer Engineering and Engineering Technology, University of Houston, Houston, TX, USA. D. Wang is with The Hong Kong Polytechnic University, Hong Kong. (e-mail: \{jshen24, jshi23@central, zhan2\}@uh.edu, dan.wang@polyu.edu.hk).
}}
\markboth{IEEE Power Energy Society General Meeting, 2026}{}
\maketitle

\vspace{-9em}

\maketitle

\begin{abstract}
The European Union's Carbon Border Adjustment Mechanism (CBAM) creates a complex challenge for the interconnected European electricity market. Traditional static analyses often miss the cross-border spillover effects that are vital for understanding this policy. This paper addresses this gap by developing a spatio-temporal Graph Neural Network (GNN) framework. It quantifies how CBAM affects electricity prices and carbon intensity (CI) at the same time. We modeled a subgraph of eight European countries. Our results suggest that CBAM is not just a uniform tax. Instead, it acts as a tool that transforms the market and creates structural differences. In our simulated scenarios, we observe that low-carbon countries like France and Switzerland can gain a competitive advantage. This suggests a potential decrease in their domestic electricity prices. Meanwhile, high-carbon countries like Poland face a double burden of rising costs. We identify the primary driver as a fundamental shift in the market's merit order.

\end{abstract}

\begin{IEEEkeywords}
Carbon Border Adjustment Mechanism (CBAM), European Electricity Market, Graph Neural Network (GNN), Network Analysis, Electricity Prices, Carbon Intensity.
\end{IEEEkeywords}

\IEEEpeerreviewmaketitle
\section{Introduction}

The European Union (EU) has committed to ambitious climate goals under the European Green Deal \cite{eu_green_deal}. This includes the Carbon Border Adjustment Mechanism (CBAM) to address carbon leakage \cite{boehringer_carbon_leakage, cbam_regulation}. This policy arrives while people are very sensitive about European energy costs. This follows a recent energy crisis that caused prices to surge. Consequently, there is significant public concern that CBAM will be just another tax that burdens consumers.

This paper challenges that assumption. We ask a simple question. Is this blame accurate? Does CBAM add costs uniformly, or could its true impact on electricity prices be more complex and asymmetric? The question is not trivial because the European power grid is a deeply interconnected network. It is not a set of isolated markets \cite{eurelectric_cbam, eu_market_integration}. Static, single-country analyses often miss the dynamic, cross-border spillover effects that define this system. To address this gap, we develop a modeling framework based on Graph Neural Networks (GNNs). This allows us to capture these dynamic, networked effects better than static analysis.

Our analysis suggests that CBAM is not a uniform financial burden. Instead, it acts as a market transformation tool that creates clear differences between countries. We find that low-carbon countries like France and Switzerland gain a significant competitive advantage. In our simulated scenarios, the model suggests that CBAM may lead to a decrease in domestic electricity prices for these nations. Conversely, high-carbon grids like Poland and the Czech Republic face a dual challenge of rising costs and lost competitiveness. We identify the primary driver as a fundamental shift of the market's merit order. The main contributions of this paper are threefold:
\begin{itemize}
    \item We develop a novel dual-target GNN framework to model the relationship between price and carbon intensity. This extends beyond previous single-task forecasting models \cite{zhang_gnn_2023}.
    \item We provide quantitative, network-aware evidence that challenges the idea of a uniform tax. We demonstrate asymmetric economic impacts in our simulations, including potential price decreases for specific regions.
    \item We identify the market mechanisms like merit-order shifting that drive this divergence. Because we cannot observe ground truth for a policy that has not fully happened yet, we focus on internal validity. We support our findings through predictive back-testing on observed data, robustness checks, and placebo tests.
\end{itemize}
\section{Background and Methodology}

\subsection{CBAM and the Network-Based Challenge}

The CBAM regulation requires importers to submit certificates for the embedded emissions of imported electricity \cite{cbam_regulation}. The European market is highly coupled and managed by ENTSO-E \cite{entsoe_role}. In this system, electricity is pooled. This makes it nearly impossible to trace the exact origin and carbon content of an imported megawatt-hour \cite{eurelectric_cbam}. This complexity requires a network-based model. A tax applied at one border like Germany will not stop there. Its cost will factor into market bids. This influences trade and alters supply and demand for all neighbors like France and Poland. It eventually creates spillover effects \cite{adekoya_spillover}. These effects vary over time. This requires a graph analysis that considers both space and time.

\subsection{Spatiotemporal GNN Framework}
Our approach extends the carbon intensity forecasting model (CFCG) presented in \cite{zhang_gnn_2023}. The original CFCG model is a single-task framework. It was designed only to predict carbon intensity (CI). Our main innovation is a dual-target, multi-task learning architecture. This allows us to model and predict two coupled market variables together: electricity Price ($Y_{Price}$) and Carbon Intensity ($Y_{CI}$). This section details the theoretical foundation of our framework.

\subsubsection{Dual-Target Problem Formulation}
Price and CI are not independent. Under CBAM, they are explicitly linked. The policy adds a carbon cost to the base price: $P_t^{CBAM} = P_t^{base} + \tau \cdot CI_t + \epsilon_t$. This creates a non-zero covariance ($\text{Cov}(CI_t, P_t) \neq 0$) that a single-task model would miss. Therefore, we formulate the problem as a multi-task learning challenge.

The original CFCG model optimized a single loss $\mathcal{L}_{CFCG} = \min_{\theta} \mathbb{E}[\ell(f_\theta(x), y_{CI})]$. In contrast, our dual-target framework optimizes a composite loss function:
\begin{equation}
\small
\begin{split}
\mathcal{L}_{dual} = \min_{\theta, \phi}&\mathbb{E}_{(x,y) \sim \mathcal{D}} \Big[  \lambda_1 \cdot \text{MSE}(Y_{CI}, \hat{Y}_{CI}) \\
&+ \lambda_2 \cdot \text{MSE}(Y_{Price}, \hat{Y}_{Price})  + \lambda_3 \cdot \mathcal{L}_{Corr} \Big]
\end{split}
\label{eq:dual_loss}
\end{equation}
Here, $\lambda_1, \lambda_2, \lambda_3$ are weighting hyperparameters. The first two terms are standard Mean Squared Error (MSE) losses for each task. The third term, $\mathcal{L}_{Corr}$, is a novel Correlation Loss. It ensures the model preserves the underlying correlation structure between the two targets:
\begin{equation}
\small
    \mathcal{L}_{Corr} = \left|\rho(Y_{CI}, Y_{Price}) - \rho(\hat{Y}_{CI}, \hat{Y}_{Price})\right|
    \label{eq:corr_loss}
\end{equation}
where $\rho(\cdot, \cdot)$ is the Pearson correlation coefficient. This term acts as a coupling regularization. This approach is based on multi-task learning theory \cite{caruana_multitask, ruder_multitask}, which provides two key benefits: \textit{Inductive Bias Transfer}, where the model is forced to find a joint hypothesis space $\mathcal{H}_{joint} = \mathcal{H}_{CI} \cap \mathcal{H}_{Price}$ leading to more generalizable features; and \textit{Statistical Efficiency}, as joint estimation reduces parameter variance compared to independent training.

\subsubsection{Model Architecture and Optimization}
\begin{figure}[t]
    \centering

    \includegraphics[width=0.4\textwidth]{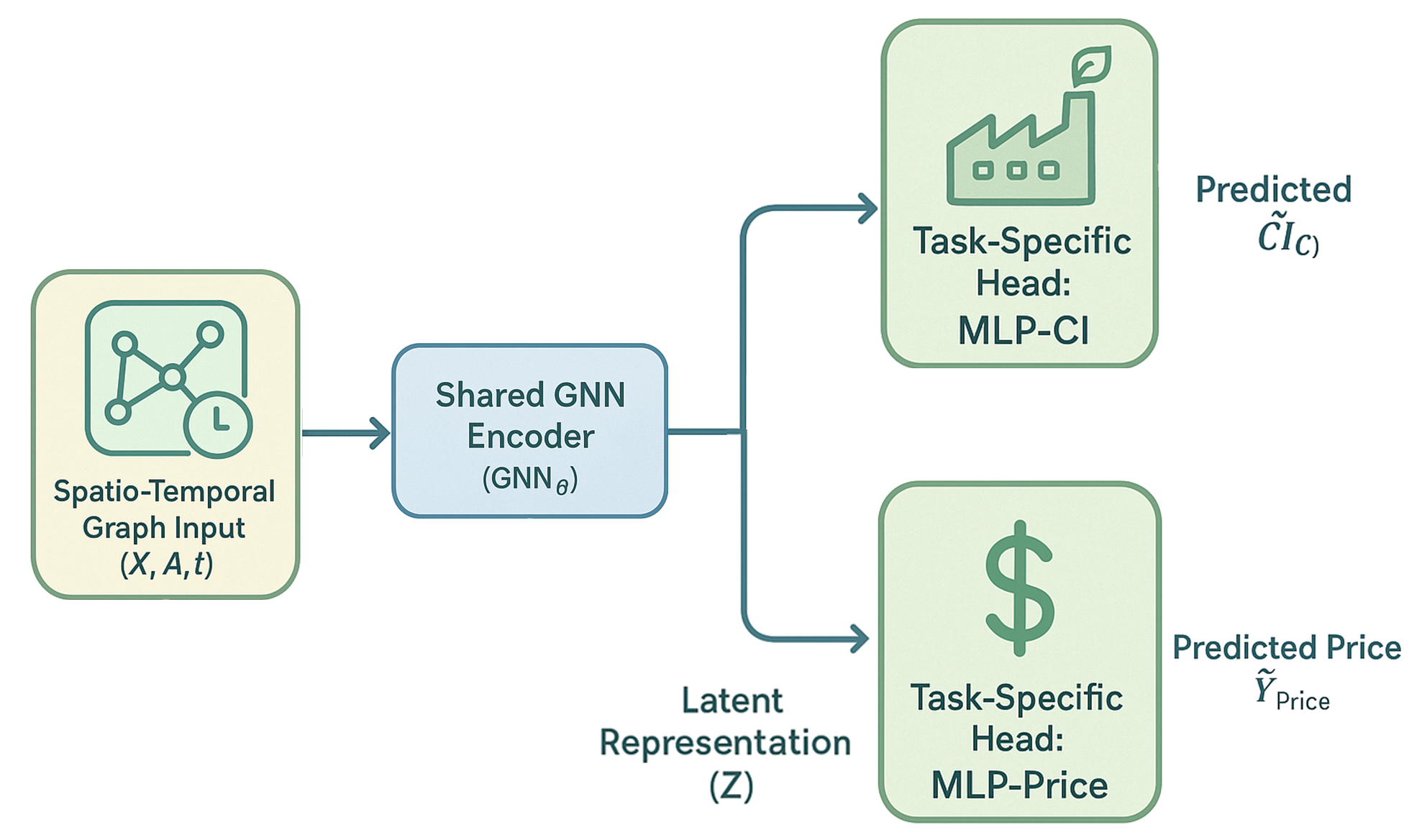}
    \caption{Our dual-target GNN architecture. It shows the shared encoder and parallel task-specific (CI and Price) MLP heads.}
    \label{fig:model_arch_placeholder}
    \vspace{-5mm}
\end{figure}

We design a parallel architecture to learn this representation efficiently (Fig. \ref{fig:model_arch_placeholder}). Instead of using the complex GNN-LSTM stacks found in the original CFCG, our model employs a shared GNN encoder ($\text{GNN}_\theta$) to extract features from the input graph $G=(V, E)$, utilizing node features $X$ and adjacency matrix $A$. This shared encoder learns a latent representation $Z$ that captures the network-wide state:
\vspace{-2mm}
\begin{equation}
\small
    Z = \text{GNN}_\theta(X, A) \in \mathbb{R}^{|V| \times d}
    \label{eq:shared_encoder}
    \vspace{-2mm}
\end{equation}
Subsequently, this shared representation $Z$ is processed by two parallel, independent Task-Specific Heads implemented as Multi-Layer Perceptrons (MLPs). These simpler heads correspond to our dual targets and generate the final predictions:
\vspace{-3mm}
\begin{equation}
\small
    \hat{Y}_{CI} = \text{MLP}_{CI}(Z), \quad \hat{Y}_{Price} = \text{MLP}_{Price}(Z)
    \label{eq:mlp_heads}
    \vspace{-2mm}
\end{equation}

\begin{algorithm}[t]\small 
\caption{Dual-Target GNN Training Framework}
\label{alg:framework}
\begin{algorithmic}[1]
\State \textbf{Input:} Graph $G=(V, E)$, Features $X$ (incl. policy), Adjacency $A$, Ground Truth $Y_{CI}, Y_{Price}$, Loss weights $\lambda_1, \lambda_2, \lambda_3$
\State \textbf{Output:} Trained models $\text{GNN}_\theta$, $\text{MLP}_{CI}$, $\text{MLP}_{Price}$
\State Initialize parameters $\theta$ (for $\text{GNN}_\theta$), $\phi_{CI}$ (for $\text{MLP}_{CI}$), $\phi_{Price}$ (for $\text{MLP}_{Price}$)
\For{each training epoch}
    \State \Comment{Shared GNN Encoder (Eq. \ref{eq:shared_encoder})}
    \State $Z \gets \text{GNN}_\theta(X, A)$ \Comment{Task-specific attention applied}
    
    \State \Comment{Parallel Task-Specific Heads (Eq. \ref{eq:mlp_heads})}
    \State $\hat{Y}_{CI} \gets \text{MLP}_{CI}(Z; \phi_{CI})$
    \State $\hat{Y}_{Price} \gets \text{MLP}_{Price}(Z; \phi_{Price})$
    
    \State \Comment{Calculate Multi-Task Loss (Eq. \ref{eq:dual_loss})}
    \State $\mathcal{L}_{MSE-CI} \gets \text{MSE}(Y_{CI}, \hat{Y}_{CI})$
    \State $\mathcal{L}_{MSE-Price} \gets \text{MSE}(Y_{Price}, \hat{Y}_{Price})$
    \State $\mathcal{L}_{Corr} \gets \left|\rho(Y_{CI}, Y_{Price}) - \rho(\hat{Y}_{CI}, \hat{Y}_{Price})\right|$
    \State $\mathcal{L}_{dual} \gets \lambda_1 \mathcal{L}_{MSE-CI} + \lambda_2 \mathcal{L}_{MSE-Price} + \lambda_3 \mathcal{L}_{Corr}$
    
    \State \Comment{Backpropagation}
    \State Update $\theta, \phi_{CI}, \phi_{Price}$ using $\nabla \mathcal{L}_{dual}$
\EndFor
\State \textbf{return} $\theta, \phi_{CI}, \phi_{Price}$
\end{algorithmic}
\end{algorithm}

This architecture is much more efficient. The complexity of the original GNN in CFCG is $\mathcal{O}_{GNN} = \mathcal{O}(|V| \cdot d_{in} \cdot d_{out} \cdot |R|)$. Here, $|R|$ is the number of relations. Our MLP-based approach is $\mathcal{O}_{MLP} = \mathcal{O}(|V| \cdot d_{in} \cdot h + h \cdot d_{out})$, where $h$ is much smaller than $d_{in} \cdot |R|$. This simplification is supported by the Universal Approximation Theorem for GNNs. It states that GNNs can approximate any continuous function on a compact set of graphs \cite{xu_gnn_approx}. Our shared encoder finds a $Z$ that maximizes the mutual information for both tasks. This ensures minimal information loss.

\begin{figure*}[t]
    \centering

    \includegraphics[width=1\textwidth]{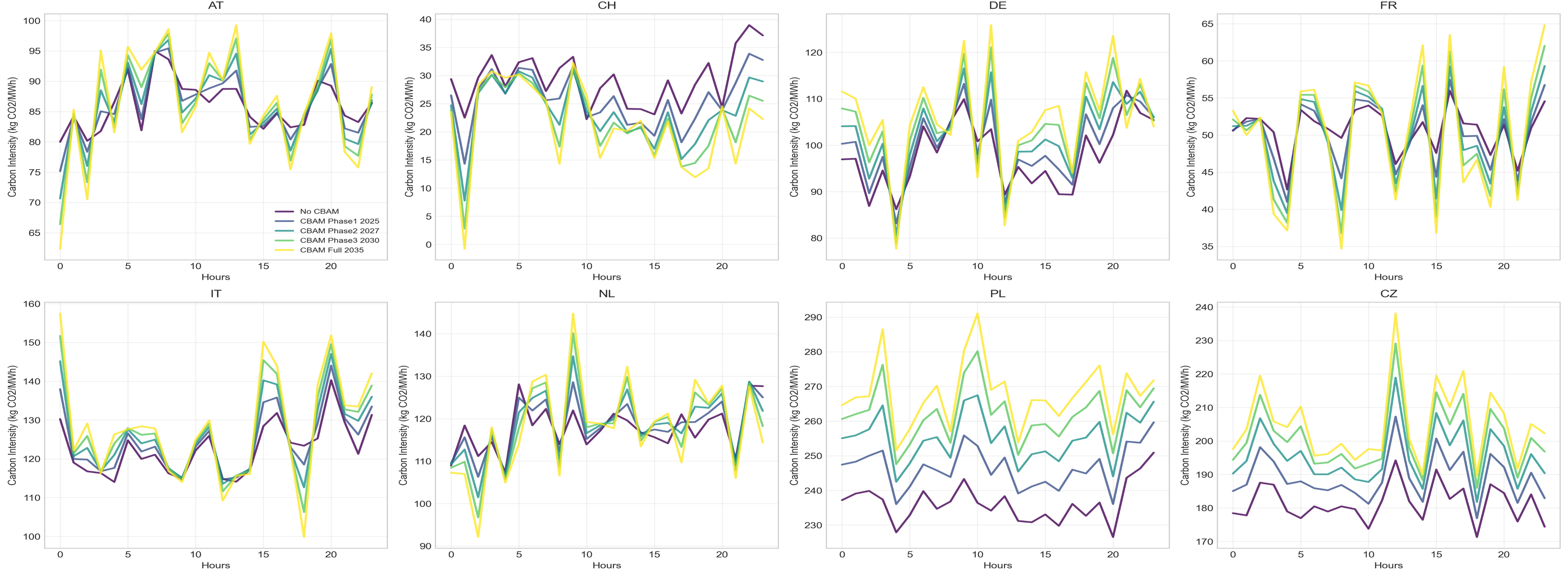} 
    \caption{Predicted CBAM impact on hourly CI. The policy suppresses CI in high-carbon nations like Poland and Czech Republic but not low-carbon ones like France and Switzerland.}
    \label{fig:ci_predictions}
    \vspace{-5mm}
\end{figure*}

\subsubsection{Task-Specific Attention Enhancement}
We enhance the GNN encoder with task-specific attention mechanisms to refine the shared representation. The shared encoder learns a general graph structure. However, the attention weights for predicting CI may differ from those for predicting Price. We implement this by making the attention calculation dependent on the task:
\vspace{-2mm}
\begin{equation}
\small
\resizebox{0.9\columnwidth}{!}{$
    \alpha_{ij}^{task} = \frac{\exp(\text{LeakyReLU}(a_{task}^T [W_{task} h_i \| W_{task} h_j]))}{\sum_{k \in \mathcal{N}(i)} \exp(\text{LeakyReLU}(a_{task}^T [W_{task} h_i \| W_{task} h_k]))}
$}
\label{eq:attention}
\end{equation}
where $task \in \{CI, Price\}$. This allows the model to weigh neighbors ($j$) differently when creating the representation for a node ($i$). It depends on the final objective and improves how the parallel MLP heads specialize.
\vspace{-1mm}
\section{Experimental Setup}

\begin{figure*}[t]
    \centering
    \includegraphics[width=1\textwidth]{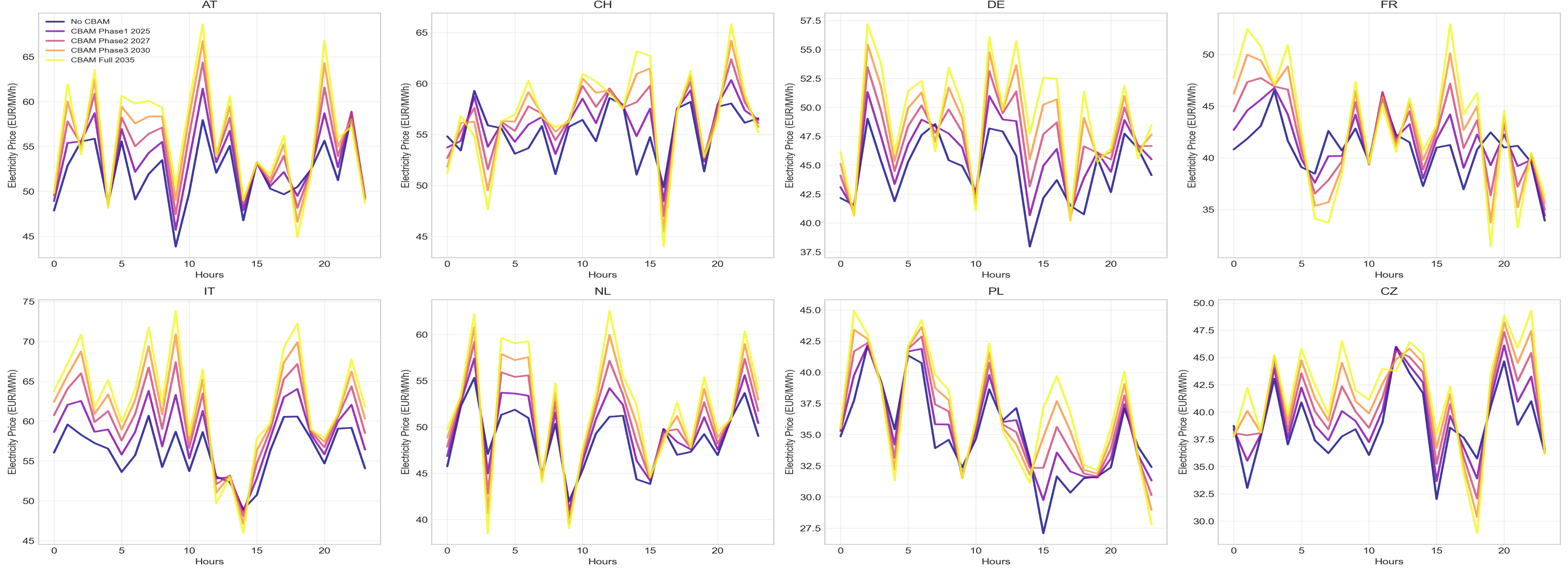} 
    \caption{Predicted CBAM impact on hourly Electricity Prices. Note the counter-intuitive price decrease in low-carbon nations (CH, NL).}
    \label{fig:price_predictions}
    \vspace{-5mm}
\end{figure*}

\subsection{Data and Scope}
Our analysis focuses on a critical subgraph of the interconnected European grid: Austria (AT), Switzerland (CH), Germany (DE), France (FR), Italy (IT), Netherlands (NL), Poland (PL), and the Czech Republic (CZ). We utilize public, hourly time-series data from the ENTSO-E Transparency Platform \cite{entsoe_data} (8,759 observations), covering demand, generation mix, cross-border flows, and day-ahead prices. The model is trained on 70\% of the data, validated on 15\%, and tested on the remaining 15\%.

To facilitate our analysis, we categorize these countries based on their average generation-based CI: Low-Carbon (CH, FR; CI $<$ 50 kg CO2/MWh), Medium-Carbon (AT, DE, IT, NL; 50 $<$ CI $<$ 130), and High-Carbon (PL, CZ; CI $>$ 130) \cite{ember_review_2023}.

\subsection{CBAM Scenario Design}
We simulate five scenarios with linear CBAM intensity increments (0\%, 25\%, 50\%, 75\%, 100\%) targeting implementation years 2025--2035.
The financial impact of the policy is calculated as an added cost to cross-border transactions, based on the carbon intensity of the exporting grid. A key assumption of our model is the establishment of a carbon intensity threshold, below which the CBAM cost is not applied. Based on the low-carbon countries (CH and FR) identified in our preliminary analysis, we set this threshold at 50 kg CO2/MWh. We also assume a proxy EU Emissions Trading System (ETS) price of 85 EUR/tCO2, a reasonable estimate reflecting market volatility and price levels seen in 2023-2024 \cite{iea_ets_price}.

The CBAM cost is thus formulated as:
\begin{equation}
\small
    Cost \text{ (EUR/MWh)} = \max(0, CI - 50) \times Intensity \times \frac{ETS}{1000}
    \label{eq:cbam_cost}
\end{equation}
where $CI$ is the hourly carbon intensity of the exporter (kg CO2/MWh), $ETS$ is the carbon price (EUR/tCO2), and $Intensity$ is the implementation phase (0.25 to 1.0). Section IV reports sensitivity over CI threshold and ETS price, and a placebo policy injection test. This calculated $Cost$ is the core policy variable injected into our model to predict the market's response.
\vspace{-3mm}
\section{Results and Analysis}

Our analytical framework allows us to analyze the differential impacts of the CBAM policy across all countries in the network, both in terms of environmental effectiveness (carbon intensity) and economic consequences (electricity price). 
Before detailing these impacts, we verified the model's reliability: our proposed framework achieves a test RMSE of 4.20 EUR/MWh for price prediction, outperforming the baseline spatial-lag model (RMSE: 5.42) and ablations lacking policy features (see Table \ref{tab:robustness}), thereby establishing a reliable foundation for counterfactual inference.

\subsection{Differential Impact on Carbon Intensity and Price}

Our primary finding (Fig. \ref{fig:ci_predictions}) is that CBAM creates divergent results. For PL and CZ, the model predicts a significant CI reduction as policy intensity increases, while the impact on low-CI nations (FR, CH) is negligible. 

More strikingly, Fig. \ref{fig:price_predictions} reveals a counter-intuitive finding: while high-carbon countries (PL, CZ) experience a clear price increase, low-carbon countries (CH, NL) show systematically lower prices under a Full CBAM scenario. This suggests CBAM is not a simple tax but a market-restructuring tool.

\begin{figure}[t]
    \centering
    \includegraphics[width=0.85\columnwidth]{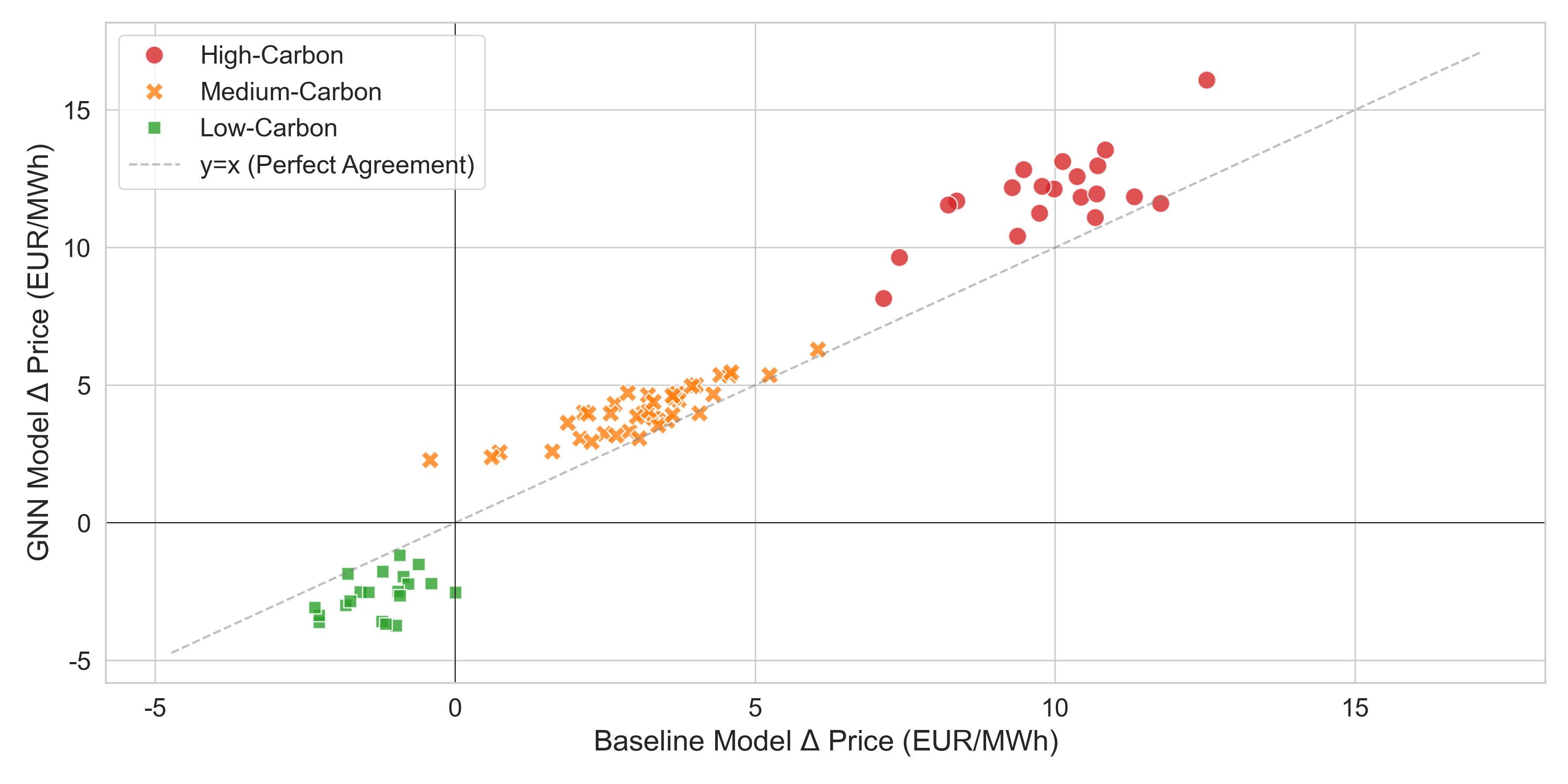}
    \caption{Scatter plot comparing price impacts predicted by our GNN (y-axis) vs. a traditional spatial-lag baseline (x-axis). }
    \label{fig:baseline_comparison}
    \vspace{-5mm}
\end{figure}

To further validate this asymmetric impact, we benchmarked our GNN model against a traditional spatial-lag econometric baseline. As shown in Fig. \ref{fig:baseline_comparison}, the two models exhibit strong directional agreement, with high-carbon nations consistently clustering in the positive impact quadrant and low-carbon nations in the negative zone. This methodological consensus confirms that the observed price asymmetry is a robust structural feature of the European grid, rather than a model-specific artifact.

\vspace{-3mm}
\subsection{Geographic Divergence of Impacts}

\begin{figure}[t]
    \centering
    \includegraphics[width=0.5\textwidth]{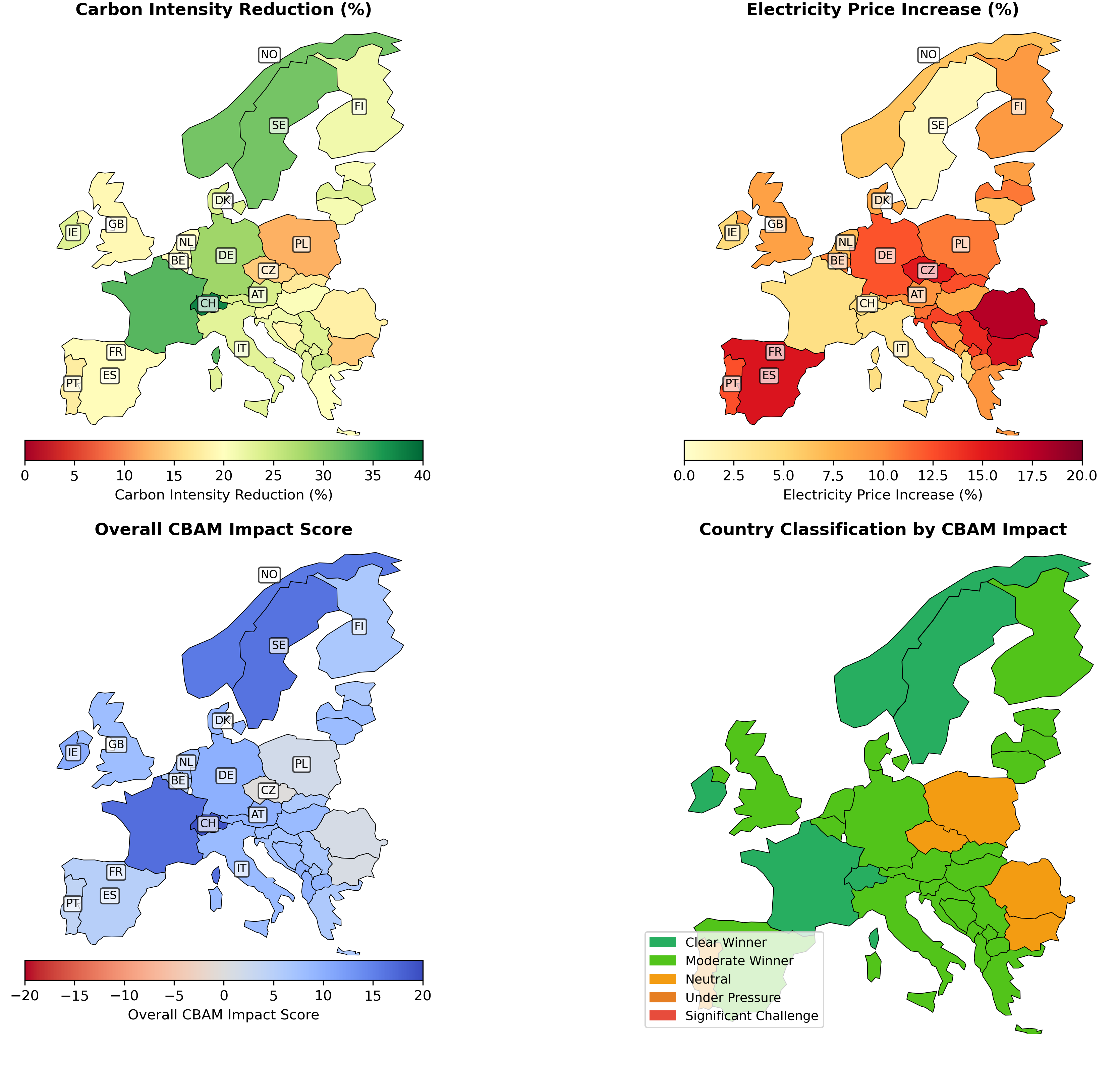}
    \caption{Geographic distribution of CBAM's impact, identifying regions with favorable outcomes (e.g., CH, FR) and those facing significant challenges (e.g., PL, CZ).}
    \label{fig:impact_maps}
    \vspace{-5mm}
\end{figure}

The national-level price and CI impacts aggregate into a clear geographic pattern of asymmetric impacts, as visualized in Fig. \ref{fig:impact_maps}. The Carbon Intensity Reduction map confirms the environmental gains are concentrated in the high-CI regions of Eastern Europe. The Price Increase map illustrates the economic trade-off, showing that these same regions bear the brunt of the costs. Our impact classification map synthesizes the net effect: regions with favorable outcomes (CH, FR, and SE); nations facing significant challenges (PL, CZ); and transitioning hubs (DE and AT).

\subsection{Uncovering the Mechanisms of Transformation}
The key question arising from these results is: why do prices decrease for clean-energy countries? Our analysis points to two fundamental market mechanisms. 

\begin{figure}[t]
    \centering
    \includegraphics[width=0.40\textwidth]{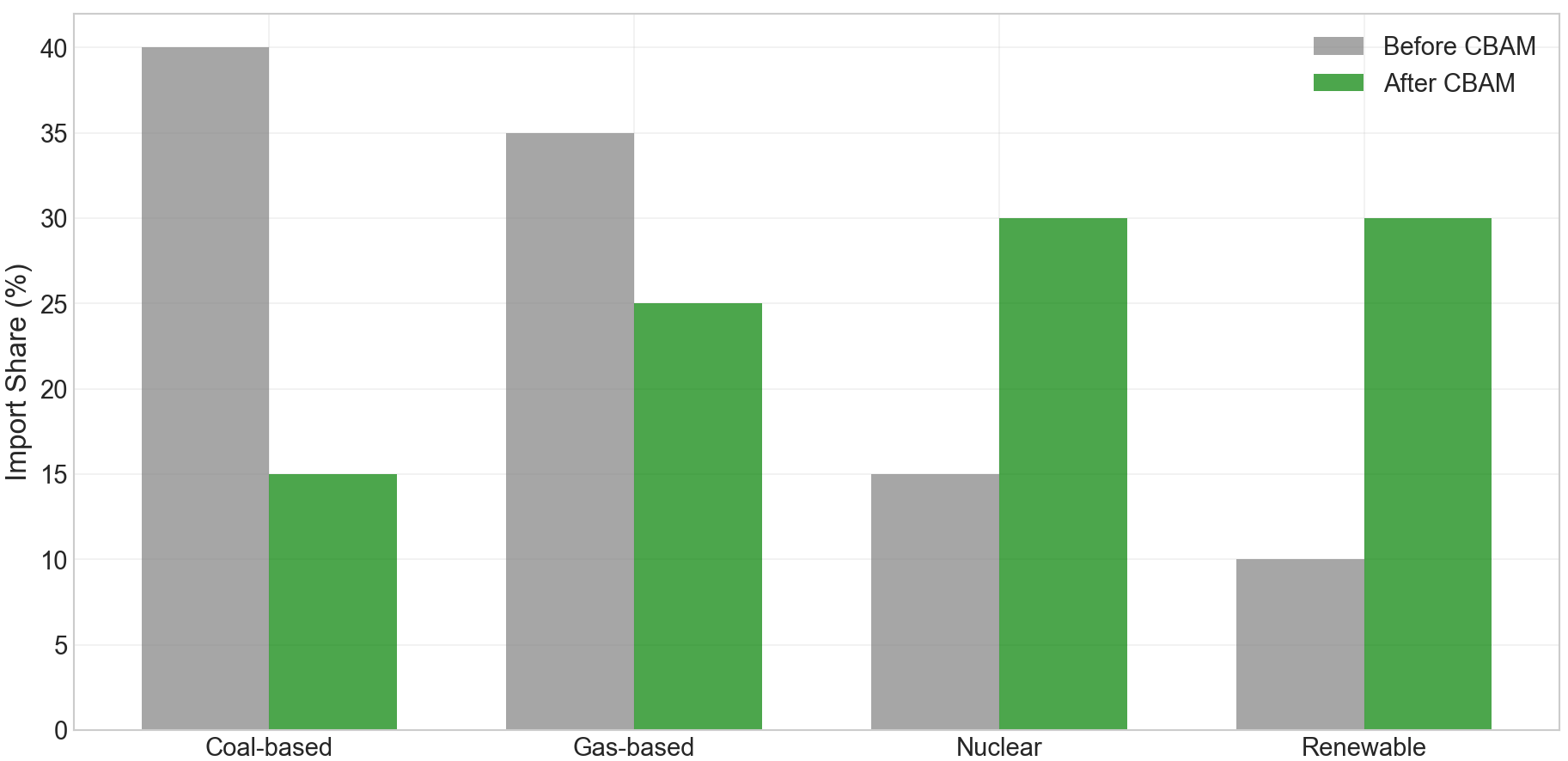} 
    \caption{Import Source Substitution under Full CBAM, showing a structural shift from coal to nuclear and renewables.}
    \label{fig:import_substitution}
    \vspace{-5mm}
\end{figure}

First, CBAM triggers a dramatic Import Source Substitution (Fig. \ref{fig:import_substitution}), where market reliance on coal-based imports plummets (from 40\% to 15\%), absorbed primarily by nuclear (15\% to 30\%) and renewables (10\% to 30\%). 

\begin{figure}[t]
    \centering
    \includegraphics[width=0.85\columnwidth]{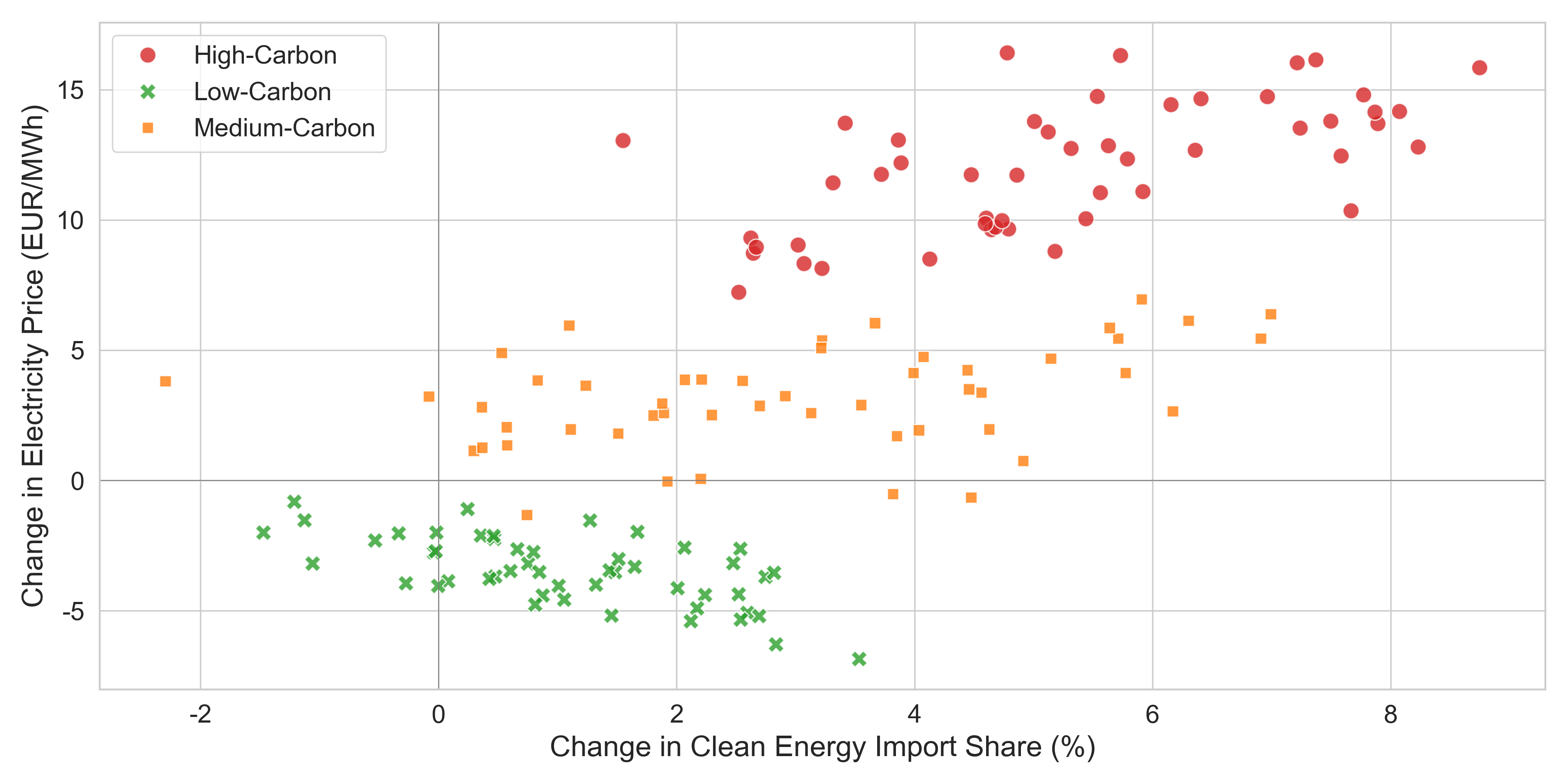}
    \caption{The trade-off between grid decarbonization (x-axis) and economic cost (y-axis).}
    \label{fig:mechanism}
    \vspace{-5mm}
\end{figure}

We quantify this substitution mechanism in Fig. \ref{fig:mechanism}, which correlates the change in clean electricity import share with domestic price impacts. The scatter plot reveals distinct clusters: High-carbon nations (red) pay a premium for substitution (driving prices up), whereas low-carbon nations (green) benefit from improved trade terms.

\begin{figure}[t]
    \centering
    \includegraphics[width=0.41\textwidth]{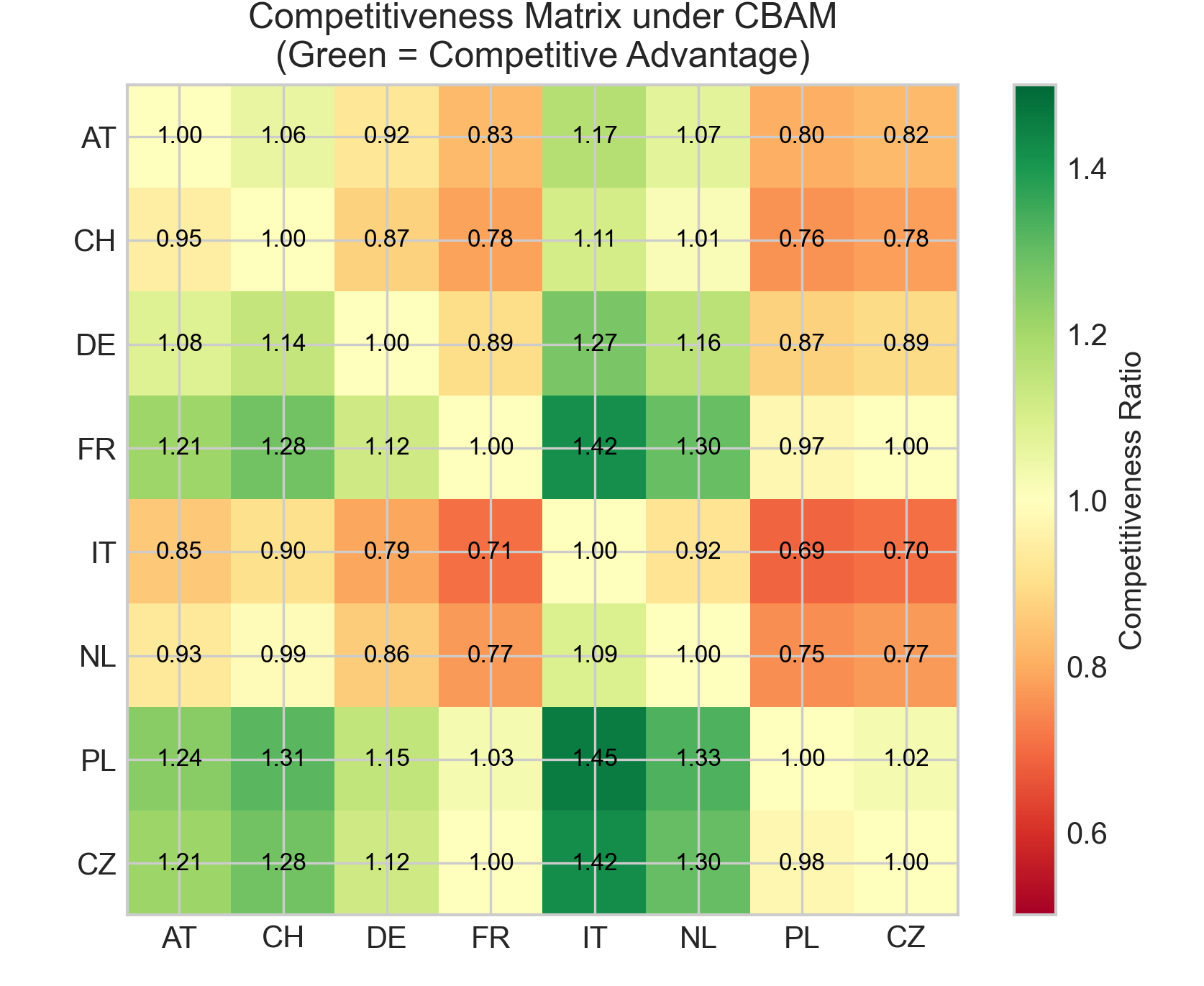}
    \caption{CBAM Competitiveness Matrix. Low-carbon nations (CH, FR) gain a competitive advantage (green) over high-carbon ones (PL, CZ).}
    \label{fig:competitiveness_matrix}
\end{figure}

Second, this substitution is driven by a Merit-Order Reshuffling. By adding a cost (Eq. 6) to high-carbon electricity, CBAM pushes fossil-fuel generators to the back of the supply queue, elevating low-cost nuclear and renewable power to a more dominant, price-setting position \cite{merit_order_renewables}. Fig. \ref{fig:competitiveness_matrix} quantifies this, showing low-carbon countries (CH, FR) gaining a strong competitive advantage (green) over high-carbon ones (PL, CZ).

\vspace{-3mm}
\subsection{Robustness and Credibility}
\vspace{-2mm}
Given the counterfactual nature of this study, verifying the causality is crucial. We conducted extensive robustness checks, summarized in Table \ref{tab:robustness}. The results confirm high directional consistency (SignAgree $>0.9$) across sensitivity sweeps. Crucially, in placebo tests where the policy variable is randomized, the effect magnitude attenuates by over 80\% (Atten. $< 0.25$), confirming that the observed impacts are causal consequences of the CBAM signal and not statistical artifacts.

\begin{table}[t]
\centering
\caption{Robustness checks. Ref: $T=50$, ETS=85. Metrics: Sign Agreement (SignAgree), Rank Correlation ($\rho$), and Attenuation Ratio (Atten.).}
\label{tab:robustness}
\scriptsize
\setlength{\tabcolsep}{3pt}
\renewcommand{\arraystretch}{1.1}
\begin{tabular}{p{1.8cm} p{2.6cm} c c c}
\toprule
\textbf{Check} & \textbf{Design} & \textbf{SignAgree} $\uparrow$ & $\boldsymbol{\rho}$ $\uparrow$ & \textbf{Atten.} $\downarrow$ \\
\midrule
Sensitivity (T)   & Threshold $\in\{25,50,75\}$ & 0.92 & 0.88 & --  \\
Sensitivity (E) & ETS $\in\{70,85,100\}$ & 0.96 & 0.95 & --  \\
Placebo (Time)    & Shuffle cost over time & 0.53 & 0.04 & 0.18 \\
Placebo (Node) & Permute cost over nodes & 0.48 & -0.12 & 0.22 \\
Baseline  & Spatial-lag Panel & 0.85 & 0.78 & -- \\
\bottomrule
\end{tabular}
\end{table}

\vspace{-0.5\baselineskip} 
\section{Policy Implications and Conclusion}

\subsection{Policy Implications}
Our findings demonstrate that CBAM functions not merely as a border tax, but as a mechanism for deep market restructuring with highly asymmetric outcomes. The statistically significant ``dual burden'' identified for high-carbon nations (e.g., PL, CZ)—characterized by unavoidable price hikes and the high cost of import substitution—underscores the critical necessity of a robust ``Just Transition'' mechanism \cite{just_transition_fund}. Policy interventions should consider using CBAM revenues not just for general redistribution, but specifically to subsidize grid modernization in eastern regions to lower the substitution premium. Conversely, the structural advantage identified for low-carbon producers (e.g., FR, CH) confirms that early decarbonizers are effectively shielded from external carbon costs, creating a powerful market-based incentive for rapid green transition.
\vspace{-2mm}
\subsection{Conclusion}
This paper presents a spatio-temporal, network-based analysis of the CBAM's impact on the European electricity market. By developing a dual-target GNN framework and validating it against spatial-lag baselines and extensive placebo tests, we move beyond static estimates to capture the dynamic spillover effects defining this complex system. Our primary finding is that CBAM drives a structural divergence rather than a uniform price increase. High-carbon countries face a robust economic challenge driven by merit-order reshuffling, while low-carbon countries gain a significant competitive advantage. Our model suggests this advantage creates a buffer against price hikes, with the potential to lower domestic electricity prices under specific network conditions. These results provide critical evidence that while CBAM effectively accelerates the green transition through import substitution, its asymmetric economic consequences must be actively managed to ensure political and economic stability.

\vspace{-2mm}
\bibliographystyle{IEEEtran}
\bibliography{IEEEexample}


\end{document}